\documentclass[letterpaper]{article} % DO NOT CHANGE THIS
\usepackage{aaai2026}  % DO NOT CHANGE THIS
\usepackage{times}  % DO NOT CHANGE THIS
\usepackage{helvet}  % DO NOT CHANGE THIS
\usepackage{courier}  % DO NOT CHANGE THIS
\usepackage[hyphens]{url}  % DO NOT CHANGE THIS
\usepackage{graphicx} % DO NOT CHANGE THIS
\usepackage{natbib}  % DO NOT CHANGE THIS AND DO NOT ADD ANY OPTIONS TO IT
\usepackage{caption} % DO NOT CHANGE THIS AND DO NOT ADD ANY OPTIONS TO IT
\usepackage{algorithm}
\usepackage{algorithmic}
\usepackage{multirow}

\usepackage{newfloat}
\usepackage{enumitem}
\usepackage{listings}
\usepackage{booktabs}
\usepackage{longtable}
\usepackage[T1]{fontenc}
\usepackage[utf8]{inputenc}
\nocopyright
\DeclareCaptionStyle{ruled}{labelfont=normalfont,labelsep=colon,strut=off} % DO NOT CHANGE THIS
\floatstyle{ruled}
\newfloat{listing}{tb}{lst}{}
\floatname{listing}{Listing}
\title{Vichara: Appellate Judgment Prediction and Explanation for the Indian Judicial System}
\author {
    Pavithra PM Nair,
    Preethu Rose Anish
}
\affiliations {
    TCS Research\\
    pavithra.nair@tcs.com, preethu.rose@tcs.com
}

\usepackage{bibentry}
\begin{document}

\maketitle

\begin{abstract}
In jurisdictions like India, where courts face an extensive backlog of cases, artificial intelligence offers transformative potential for legal judgment prediction. A critical subset of this backlog comprises appellate cases, which are formal decisions issued by higher courts reviewing the rulings of lower courts. To this end, we present \textbf{Vichara}, a novel framework tailored to the Indian judicial system that predicts and explains appellate judgments. Vichara processes English-language appellate case proceeding documents and decomposes them into \textit{decision points}. Decision points are discrete legal determinations that encapsulate the legal issue, deciding authority, outcome, reasoning, and temporal context. The structured representation isolates the core determinations and their context, enabling accurate predictions and interpretable explanations. Vichara's explanations follow a structured format, inspired by the IRAC (Issue-Rule-Application-Conclusion) framework and adapted for Indian legal reasoning. This enhances interpretability, allowing legal professionals to assess the soundness of predictions efficiently. We evaluate Vichara on two datasets, PredEx and the expert-annotated subset of the Indian Legal Documents Corpus (ILDC\_expert), using four large language models (GPT-4o mini, Llama-3.1-8B, Mistral-7B, Qwen2.5-7B). Vichara surpasses existing judgment prediction benchmarks on both datasets, with GPT-4o mini achieving the highest performance (F1: 81.5 on PredEx, 80.3 on ILDC\_expert), followed by Llama-3.1-8B. Human evaluation of the generated explanations across \textit{Clarity}, \textit{Linking}, and \textit{Usefulness} metrics highlights GPT-4o mini's superior interpretability.
\end{abstract}


\section{Introduction}

Legal Judgment Prediction (LJP) aims to algorithmically forecast judicial outcomes based on case texts. The integration of artificial intelligence (AI) into LJP systems presents a promising avenue for enhancing efficiency and transparency in judicial processes. This is particularly relevant in India, where the judiciary is burdened by a massive backlog of cases. According to data from the National Judicial Data Grid, as published on the Open Government Data Platform of India\footnote{\url{https://data.gov.in/resource/court-wise-number-cases-pending-various-courts-information-available-national-judicial-1}}, as of 21 March 2025, there were approximately 45.51 million cases pending before District and Subordinate Courts, 6.25 million cases before High Courts, and 81,598 cases before the Supreme Court. Altogether, more than 51 million cases remain unresolved across various levels of the judiciary. While the majority of pending cases are at the first-instance level, appellate cases, which are cases reviewed by higher courts following lower-court rulings, represent a critical subset. These cases are particularly important because they set legal precedent and ensure consistency across lower courts, making timely resolution essential \cite{breyer2006reflections}.

AI-driven systems for appellate judgment prediction (AJP) can assist in prioritizing appeals, evaluating legal reasoning, and generating interpretable explanations to support judicial decision-making. However, building effective AJP systems remains challenging due to the complexity and domain-specific nature of legal language and reasoning. Beyond predictive accuracy, AJP systems must generate interpretable explanations to ensure transparency and trustworthiness. Appellate judgments carry high-stakes consequences, and without structured, comprehensive explanations, AI-generated outputs are difficult to validate, contest, or rely upon in practice.  

We present Vichara, a framework for appellate judgment prediction and explanation tailored to the Indian judicial context. Named after the Sanskrit word for \textit{deliberation} or \textit{reasoned consideration}, Vichara reflects the analytical rigor inherent in judicial decision-making. The framework comprises six stages: rhetorical role classification, case context construction, decision point extraction, present court ruling generation, judgment prediction, and explanation generation. In the first stage, each sentence in the case proceeding document is classified according to its rhetorical role \citep{bhattacharya2019identification}, which refers to the function it serves within the legal discourse, such as stating facts, presenting arguments, citing precedents, or delivering rulings. From the sentences identified as facts, the framework then constructs the case context, capturing the core legal issue, the court deciding the appeal, the parties, and their stances. Next, the framework extracts structured decision points that encode the individual legal issues under consideration, the deciding authority, the outcome, the underlying reasoning, and any temporal information. The decision points where the deciding authority is the present court are retained to generate the present court ruling. Finally, the judgment outcome is derived by comparing this ruling with the appellant’s stance, and an explanation is produced using the case context, decision points, present court ruling, and predicted judgment outcome. Vichara currently operates exclusively on English-language case documents.
 
A central contribution of Vichara is its generation of structured explanations for predicted judgments. Rather than producing free-form text, Vichara outputs explanations in a standardized format comprising sections such as \textit{Facts of the Case}, \textit{Legal Issues Presented}, \textit{Applicable Law and Precedents}, \textit{Reasoning}, and \textit{Conclusion}. This format is inspired by the widely adopted IRAC (Issue–Rule–Application–Conclusion) framework \citep{metzler2002importance}  and adapted to reflect the organization of judicial reasoning in Indian court documents. The structured format enhances interpretability, allowing legal professionals to quickly assess the legal soundness of predictions. 

% Vichara is particularly well suited for routine legal matters that form a significant portion of the judicial backlog, such as property disputes, tenancy disagreements, and motor accident claims. In such cases, timely and reasoned decisions can have a substantial impact, and the structured format of AI-generated explanations enables legal professionals to quickly evaluate the reasoning behind predicted outcomes, facilitating faster review and decision support. 

We evaluate Vichara on two datasets: PredEx \citep{nigam2024legal} and the expert-annotated subset of the Indian Legal Documents Corpus (ILDC\_expert) \citep{malik2021ildc}. Our experiments use four large language models (LLMs): GPT-4o mini \citep{openai2024gpt4omini}, Llama-3.1-8B \citep{grattafiori2024llama}, Mistral-7B \citep{jiang2023mistral7b}, and Qwen2.5-7B \citep{qwen2024qwen2}. Each model is assessed on two axes: prediction performance and explanation quality. GPT-4o mini achieves the highest prediction performance (F1: 81.5 on PredEx, 80.3 on ILDC\_expert), followed by Llama-3.1-8B (F1: 76.7 on PredEx, 78.5 on ILDC\_expert). For explanation quality, we conduct human evaluation using three metrics: \textit{Clarity}, \textit{Linking}, and \textit{Usefulness}, with GPT-4o mini again receiving the highest ratings, followed by Mistral-7B. 

Our work makes the following key contributions:
\begin{itemize}
    \item We propose Vichara, a novel framework for appellate judgment prediction and explanation,  centered on decision point extraction from case proceedings.
    \item We introduce a structured explanation format grounded in legal reasoning conventions, enhancing interpretability and alignment with judicial logic.
    \item We conduct an empirical evaluation of Vichara across two datasets and four LLMs, demonstrating strong performance in both prediction and explanation quality. Vichara outperforms existing judgment prediction benchmarks on both datasets.
\end{itemize}

The rest of the paper is organized as follows. Section~\ref{related} discusses related work in legal judgment prediction. Section~\ref{dataset} and \ref{task} discuss the datasets used and the task formulation, respectively. Section~\ref{methodology} describes the Vichara framework in detail. Section~\ref{results} presents the results and analysis. Section~\ref{conclusions} concludes with a discussion of implications and future directions, and Section~\ref{limitations} presents the limitations of Vichara. For the sake of reproducibility, we have made the code accessible via a GitHub link\footnote{\url{https://github.com/pavithranair/Vichara}}.

\begin{figure*}[htbp]
  \centering
  \includegraphics[width=1\linewidth]{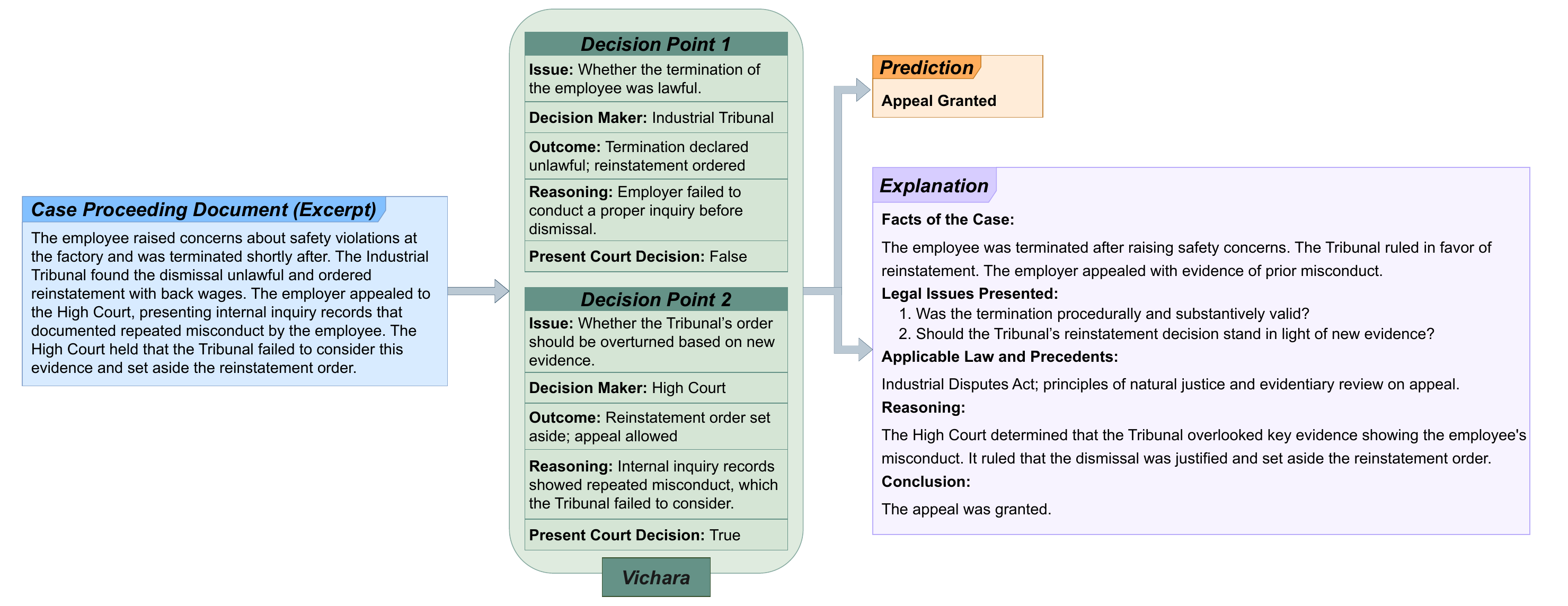} \hfill
  \caption {Vichara’s appellate judgment prediction and explanation for a sample case excerpt.}
  \label{cjpe}
\end{figure*}

\section{Related Work}
\label{related}

Legal Judgment Prediction (LJP) has advanced considerably in recent years, driven by the growing need to automate legal outcome prediction and ease the burden of increasing caseloads on judicial systems. The field has been shaped by several foundational studies that established core tasks, datasets, and modeling approaches. \citep{aletras2016predicting} first demonstrated that court decisions could be predicted from textual case descriptions using traditional feature-based models. \citep{zhong2018legal} extended this by introducing TopJudge, a multi-task framework that jointly predicts legal charges, applicable statutes, and sentence lengths, supported by the large-scale CAIL2018 dataset \citep{xiao2018cail2018}. In the European legal context, \citep{chalkidis2019neural} advanced the field with neural architectures that improved performance over prior baselines on judgment prediction tasks. \cite{sulea-etal-2017-predicting} applied text classification methods to predict case outcomes, law areas, and ruling periods for French Supreme Court cases, while analyzing the impact of temporal context and masking judges’ motivations. \cite{medvedeva2020using} demonstrate the application of large-scale statistical analysis and machine learning to European Court of Human Rights case texts to predict judicial decisions.

% The advent of LLMs has significantly expanded the scope of legal AI applications, enabling advancements in tasks such as statutory interpretation, case outcome prediction, and legal question answering. Recent studies have explored ways to make LLMs more suitable for legal contexts by enhancing their logical structure and grounding. For instance, \citep{yao2025elevating} introduced LSIM, a framework that combines reinforcement learning with retrieval-augmented generation to reduce hallucinations and improve logical consistency in legal Q\&A tasks. Similarly, \citep{zhang2025syler} proposed SyLeR, which equips LLMs with explicit syllogistic reasoning capabilities and structure-aware fine-tuning to better reflect legal logic. Despite these innovations, challenges related to faithfulness, factual grounding, and bias remain central, prompting ongoing research into provenance tracking and structured explanation techniques \citep{siino2025exploring}. 

In the Indian legal domain, research on judgment prediction has evolved rapidly, with an increasing focus on both explainability and realism in predictive settings. Foundational datasets such as ILDC \citep{malik2021ildc} and PredEx \citep{nigam2024legal} introduced large, expertly annotated collections of Indian court cases that serve as benchmarks for evaluating models on judgment prediction and explanation tasks. Fact-driven approaches have gained prominence through studies like \citep{nigam2024rethinking} and \citep{nigam2023fact}, which restrict inputs to case facts or limited procedural history, reflecting constraints faced in real-world litigation support systems. More recently, the NyayaAnumana dataset \citep{nigam2025nyayaanumana} has set a new benchmark in scale and coverage, comprising over 700,000 cases from across the Indian judiciary. Accompanied by INLegalLlama, a domain-specialized generative model, this work demonstrates substantial gains in both predictive accuracy and the coherence of generated explanations. 

% \citep{tiwari2024aalap} examine AI assistants for paralegal and legal support, showing how such systems can automate routine legal tasks. \citep{ganguly2023legal} review the application of information retrieval and natural language processing in the legal domain, outlining key challenges in handling lengthy, complex texts and limited datasets, and surveying state-of-the-art methods, tools, and resources.

Despite recent advances, most existing approaches either focus primarily on prediction, where performance still lags behind human experts, or attempt to generate explanations that fall into two categories. Some are \textit{extractive} \cite{prasad2023hierarchical, yamada2024japanese}, simply highlighting relevant case facts, while others are \textit{abstractive} \cite{nigam2024rethinking, nigam2025nyayaanumana}, producing free-form summaries. In both cases, explanations are usually unstructured and offer limited insight into how facts and legal arguments lead to the court’s decisions. In contrast, Vichara leverages LLMs to deliver strong predictive performance while producing structured, interpretable explanations that allow legal professionals to trace reasoning from evidence to outcome. By explicitly modeling the legal reasoning process, Vichara supports both practitioners and the wider public in navigating complex judicial environments, while paving the way for more transparent and accountable AI-driven legal tools.

\section{Dataset}
\label{dataset}

We evaluate our framework using two benchmark datasets from the Indian legal domain:
\begin{itemize}
    \item PredEx: PredEx \citep{nigam2024legal} is the largest publicly available source for joint judgment prediction and explanation tasks in the Indian context. We use the test split of this dataset, which contains 3,044 appellate cases from the Supreme Court of India and various high courts, annotated with binary outcome labels (Appeal Granted or Dismissed) and expert-annotated explanations.
    \item ILDC\_expert: ILDC\_expert is the expert-annotated subset of the Indian Legal Documents Corpus \citep{malik2021ildc}. It consists of 56 Supreme Court of India appellate cases, each annotated with binary outcome labels (Appeal Granted or Dismissed) and corresponding expert provided explanations.

\end{itemize}

All case proceedings and explanations in both datasets are written in the English language.

\section{Vichara Task Formulation}
\label{task}

Vichara consists of a unified pipeline with two interconnected components:

\subsection{Judgment Prediction Component}
Given a segment from a Supreme Court of India or high court appellate case proceeding document, Vichara predicts whether the court ruled in favor of the appellant. The output is a binary label: \{1, 0\}, where 1 denotes that the appeal was granted, and 0 indicates dismissal. While real cases can involve mixed outcomes, Vichara abstracts these into a single binary decision to focus on the core outcome.

\subsection{Explanation Generation Component}
Alongside predicting the outcome, Vichara produces a structured rationale explaining the prediction. The explanation captures the key facts, applicable laws, and legal reasoning, following an IRAC-inspired structure adapted to Indian legal discourse.

Figure~\ref{cjpe} illustrates Vichara's task formulation.

\section{Methodology}
\label{methodology}

\begin{figure*}[htbp]
  \centering
  \includegraphics[width=1\linewidth]{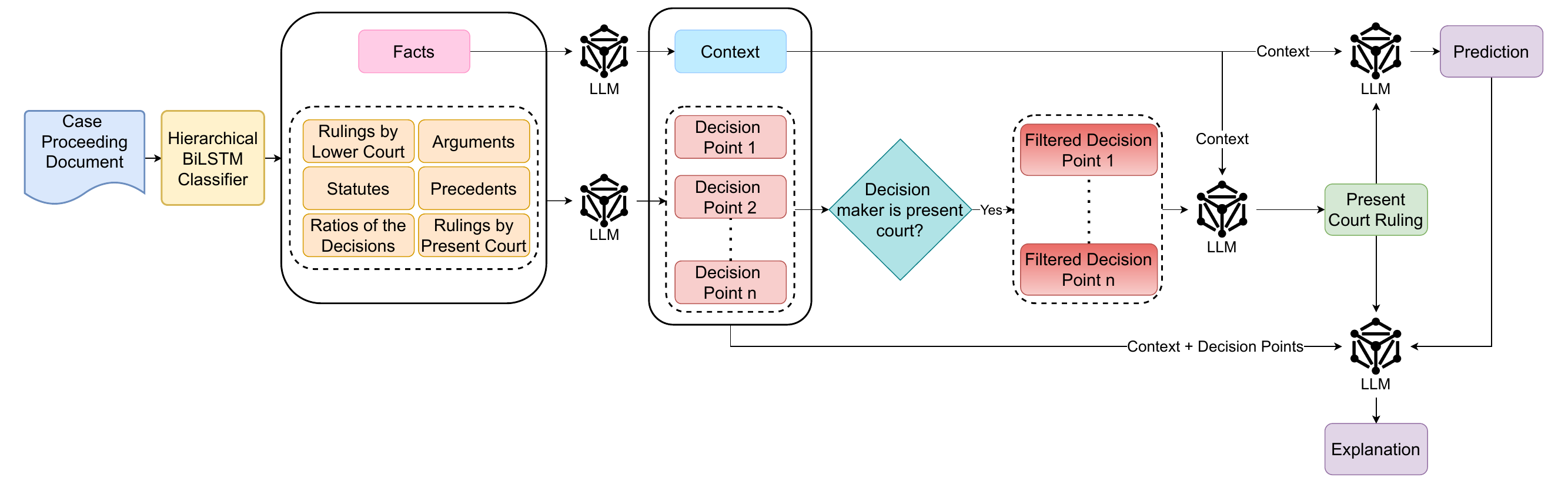} \hfill
  \caption {Vichara flow diagram}
  \label{flow}
\end{figure*}

Vichara comprises six sequential stages: rhetorical role classification, case context construction, decision point extraction, present court ruling generation, judgment prediction, and explanation generation. 

The process begins by classifying each sentence in the input case proceeding document according to its rhetorical role \citep{bhattacharya2019identification}, identifying whether it states facts, presents arguments, cites precedents, or issues rulings. All subsequent stages in the pipeline operate purely through prompting without any fine-tuning of the underlying LLMs. The case context is extracted exclusively from sentences labeled as facts, capturing the key legal issue, the court deciding the appeal, the parties, and their respective stances. Vichara extracts decision points that summarize the court’s legal determinations, filtering them to retain only those attributable to the present court. These filtered decision points are then used to generate the present court's ruling, ensuring it reflects solely the outcome of the appeal under consideration. The predicted judgment is derived by comparing this ruling with the appellant's stance. Finally, Vichara produces an IRAC-style \citep{metzler2002importance} explanation grounded in the extracted facts, laws, and reasoning. The overall architecture is illustrated in Figure~\ref{flow}, with detailed prompt templates provided in Tables~\ref{tab:prompts1}, \ref{tab:prompts2}, and \ref{tab:prompts3} in the appendix. Next, we explain each of the steps in the methodology.

\subsection{Rhetorical Role Classification}
\label{sec:rhet}
Rhetorical role labelling of sentences in a legal document refers to identifying the semantic function each sentence serves, such as stating the facts of the case, presenting arguments of the parties, or delivering the court’s judgment \citep{bhattacharya2019identification}. 

The input case proceeding document is segmented into sentences using the SAT-12L (Segment Any Text) pretrained model, fine-tuned via Low-Rank Adaptation (LoRA) for accurate sentence boundary detection in legal texts \citep{frohmann2024segment}. \citep{bhattacharya2019identification} define seven categories of rhetorical roles, namely `Facts', `Ruling by Lower Court', `Argument', `Statute', `Precedent', `Ratio of the Decision', and `Ruling by Present Court'. Each sentence in our dataset is classified into one of these seven categories by a hierarchical BiLSTM-CRF model trained to assign rhetorical roles, following the schema introduced by \citep{bhattacharya2019identification}.  

\subsection{Case Context Construction}
\label{context}

To generate case-level context, an LLM is applied to the subset of sentences labeled as `Facts' during rhetorical role classification (Section~\ref{sec:rhet}). The LLM extracts six key fields: 
\begin{itemize}[noitemsep]
    \item Appellants: The individuals or entities filing the appeal.
    \item Respondents: The opposing parties in the case.
    \item Issue: The main legal or factual question under consideration.
    \item Appellant’s Stance: The relief sought or position taken by the appellant.
    \item Respondent’s Stance: The response or objections raised by the respondent.
    \item Present Court: The court deciding the appeal. 
\end{itemize}

This structured context summarizes the parties, their positions, the present court and the core dispute in the case.  

\subsection{Decision Point Extraction}
\label{dec}

To extract decision points, the case proceeding document is divided into manageable segments for prompting. When available, explicit bullet-point lists in the document are used, which often enumerate key issues or rulings. If bullet-point structure is absent, the document is divided into 1000-token segments. This chunk size is chosen based on empirical observations that longer segments tend to degrade the quality of decision point extraction.  

These segments are passed on to an LLM to extract decision points. Decision points are discrete legal determinations made at different stages throughout the case. Each decision point is represented as a structured unit consisting of the following six fields: 
\begin{itemize}[noitemsep]
    \item Issue: The specific legal or factual question under consideration.
    \item Decision Maker: The authority that issued the decision.
    \item Outcome: The court's resolution of the issue, such as whether a claim was upheld, dismissed, or partially granted.
    \item Reasoning (optional): The rationale provided by the decision maker for the determination on that issue.
    \item Time (optional): Any explicit reference to when the decision was made, especially relevant in multi-phase cases or appeals.
    \item Present Court Decision: A Boolean flag indicating whether the decision maker is the present court deciding the appeal. 
\end{itemize}

\subsection{Present Court Ruling Generation}
\label{present}

To ensure that the LLM focuses solely on the reasoning of the present court, the extracted decision points are filtered to retain only those where the deciding authority is the present court. This prevents the framework from incorporating rulings made by lower courts, which may appear earlier in the case document and introduce confusion. We also extract the final sentence of the case document with the rhetorical role `Ruling by Present Court', referred to as the \textit{final statement}.

Using the structured case context, the filtered decision points, and the final statement, an LLM is prompted to generate a detailed summary of the present court’s ruling. The output includes specific reliefs granted or denied, any orders or directions issued, the court’s reasoning and key considerations, and relevant timelines or compliance instructions. This step captures only the outcome of the current appeal, ignoring lower court decisions or earlier findings unless explicitly referenced by the present court. 

\subsection{Judgment Prediction}
\label{pred}

To determine the binary outcome of the appeal, either granted or dismissed, an LLM compares the present court ruling (Section~\ref{present}) with the appellant’s stance extracted during context construction (Section~\ref{context}). If the ruling fully or partially grants the relief sought by the appellant, the framework outputs a prediction of 1 (Appeal Granted). Otherwise, if the relief is entirely denied, the prediction is 0 (Appeal Dismissed). 

\subsection{Structured Explanation Generation}
\label{explanation}
The final stage of the pipeline involves generating a structured explanation for the predicted outcome of the appeal. An LLM is used to synthesize the explanation. The inputs to the LLM include the case context (Section~\ref{context}), extracted decision points (Section~\ref{dec}), present court ruling (Section~\ref{present}), and the predicted outcome (Section~\ref{pred}).

The LLM is prompted to organize the explanation into five predefined sections:
\begin{itemize}[noitemsep]
    \item Facts of the Case: A brief overview of the dispute, parties involved, and procedural background.
    \item Legal Issue(s) Presented: The central legal questions that the court addressed.
    \item Applicable Law and Precedents: Statutes, constitutional provisions, or prior case law cited or relied upon.
    \item Analysis / Reasoning: The logical application of the law to the facts, incorporating relevant decision points to justify the predicted outcome.
    \item Predicted Conclusion: A summary conclusion reflecting the model’s judgment (e.g., “Appeal Granted” or “Appeal Dismissed”).
\end{itemize}

This structured format ensures that explanations are interpretable and legally grounded. 

\section{Results and Analysis}
\label{results}

In this section, we present the evaluation of Vichara on both judgment prediction and explanation quality, as well as an ablation study to assess the necessity of each stage of the Vichara framework. We evaluate the framework using four LLMs, GPT-4o mini \citep{openai2024gpt4omini}, Llama-3.1-8B\footnote{\url{https://huggingface.co/meta-llama/Llama-3.1-8B-Instruct}}, Mistral-7B\footnote{\url{https://huggingface.co/mistralai/Mistral-7B-Instruct-v0.3}} and Qwen2.5-7B\footnote{\url{https://huggingface.co/Qwen/Qwen2.5-7B-Instruct}}. All experiments were repeated across 5 independent random seeds to account for the non-deterministic nature of LLMs. Reported results for judgment prediction and automated evaluation of explanations correspond to the mean performance across seeds, with standard deviations included to indicate robustness.

\subsection{Results for Judgment Prediction}

\begin{table*}[t]
\centering
\small
\setlength{\tabcolsep}{3pt}
\begin{tabular}{lcccc|cccc}
\toprule
\multirow{2}{*}{Model} & \multicolumn{4}{c|}{\textbf{PredEx}} & \multicolumn{4}{c}{\textbf{ILDC\_expert}} \\
\cmidrule(lr){2-5} \cmidrule(lr){6-9}
 & Accuracy & \shortstack{Macro\\ Precision} & \shortstack{Macro\\ Recall} & Macro F1 & Accuracy & \shortstack{Macro\\ Precision} & \shortstack{Macro\\ Recall} & Macro F1 \\
\midrule
\textbf{GPT-4o mini} & \textbf{81.62 ± 0.41} & \textbf{81.57 ± 0.43} & \textbf{81.45 ± 0.40} & \textbf{81.50 ± 0.42} & \textbf{80.36 ± 0.37} & \textbf{81.25 ± 0.34} & \textbf{80.65 ± 0.36} & \textbf{80.30 ± 0.35} \\
Llama-3.1-8B & 76.93 ± 0.49 & 76.75 ± 0.51 & 76.60 ± 0.48 & 76.66 ± 0.50 & 78.57 ± 0.42 & 78.71 ± 0.40 & 78.42 ± 0.39 & 78.46 ± 0.41 \\
Mistral-7B & 69.41 ± 0.55 & 69.21 ± 0.53 & 69.00 ± 0.52 & 69.04 ± 0.54 & 75.00 ± 0.46 & 75.48 ± 0.44 & 75.22 ± 0.43 & 74.97 ± 0.45 \\
Qwen2.5-7B & 71.73 ± 0.50 & 70.06 ± 0.52 & 71.97 ± 0.48 & 72.10 ± 0.51 & 76.79 ± 0.44 & 78.48 ± 0.42 & 77.20 ± 0.45 & 76.60 ± 0.43 \\
\shortstack[l]{INLegalLlama \\\cite{nigam2025nyayaanumana}} & 76.05 & 76.23 & 76.05 & 76.01  & 72.23 & 73.01 & 72.23 & 71.98  \\
\bottomrule
\end{tabular}
\caption{Judgment prediction performance comparison of models on PredEx and ILDC\_expert datasets, reported as mean ± standard deviation across 5 random seeds. The LLMs listed (GPT-4o mini, Llama-3.1-8B, Mistral-7B, Qwen2.5-7B) are used within the Vichara framework. The best-performing model for each dataset is highlighted in bold. We have also included the state-of-the-art baseline, INLegalLlama \cite{nigam2025nyayaanumana}, for comparison.}
\label{tab:pred}
\end{table*}

We evaluate the binary outcome of judgment prediction using standard classification metrics: accuracy, precision, recall, and F1-score. To ensure balanced evaluation across both classes (Appeal Granted and Dismissed), we report macro-averaged scores. Results are presented in Table~\ref{tab:pred}. 

Among the LLMs evaluated in our experiments within the Vichara framework, GPT-4o mini achieves the highest scores on both datasets, followed by Llama-3.1-8B, Qwen2.5-7B, and Mistral-7B. Compared to the state-of-the-art baseline, INLegalLlama \cite{nigam2025nyayaanumana}, all LLMs evaluated in our experiments surpass its performance on the ILDC\_expert dataset, while on PredEx, GPT-4o mini and Llama-3.1-8B outperform INLegalLlama.

\subsection{Results for Explanation}
\label{results:ex}
\setlength{\tabcolsep}{5pt}
\begin{table}[t]
\centering
\begin{tabular}{lccc}
\hline
 & \textbf{Clarity} & \textbf{Linking} & \textbf{Usefulness} \\ 
\hline
\textbf{GPT-4o mini} & \textbf{4.57 ± 0.22} & \textbf{4.96 ± 0.21} & \textbf{4.37 ± 0.18} \\
Llama-3.1-8B & 3.43 ± 0.30 & 3.77 ± 0.25 & 3.29 ± 0.28 \\
Mistral-7B & 4.11 ± 0.27 & 4.44 ± 0.22 & 3.85 ± 0.24 \\
Qwen2.5-7B & 3.44 ± 0.26 & 3.29 ± 0.31 & 3.33 ± 0.29 \\ 
\hline
\end{tabular}
\caption{Expert evaluation results for the explanation task. Values are reported as mean ± standard deviation across three evaluators. Bold values indicate the highest score for each metric.}
\label{tab:exp}
\end{table}

\begin{table*}[t]
\small
\centering
\setlength{\tabcolsep}{4pt}
\renewcommand{\arraystretch}{1.2}
\begin{tabular}{lcccccccc}
\hline
\textbf{Models} & \multicolumn{6}{c}{\textbf{Lexical Based Evaluation (\%)}} & \multicolumn{2}{c}{\textbf{Semantic Evaluation (\%)}} \\
\cline{2-7} \cline{8-9}
& \textbf{Rouge-1} & \textbf{Rouge-2} & \textbf{Rouge-L} & \textbf{BLEU} & \textbf{METEOR} & & \textbf{BERTScore(F1)} & \textbf{BLANC} \\
\hline
\multicolumn{9}{c}{\textbf{PredEx}} \\
\hline
GPT-4o mini & 38.60 ± 0.42 & 14.52 ± 0.33 & 19.35 ± 0.29 & 3.24 ± 0.15 & 21.87 ± 0.31 & & 82.79 ± 0.38 & 13.48 ± 0.21 \\
Llama-3.1-8B & 32.14 ± 0.40 & 14.10 ± 0.30 & 17.49 ± 0.25 & 3.06 ± 0.12 & 19.06 ± 0.28 & & 81.55 ± 0.36 & 12.44 ± 0.20 \\
Mistral-7B & 35.75 ± 0.41 & 15.53 ± 0.32 & 18.80 ± 0.27 & 2.74 ± 0.13 & 20.61 ± 0.30 & & \textbf{82.90 ± 0.37} & 13.41 ± 0.22 \\
\textbf{Qwen2.5-7B} & \textbf{38.70 ± 0.43} & \textbf{15.80 ± 0.34} & \textbf{19.49 ± 0.30} & \textbf{7.06 ± 0.25} & \textbf{23.77 ± 0.35} & & 81.66 ± 0.38 & \textbf{13.59 ± 0.23} \\
\hline
\multicolumn{9}{c}{\textbf{ILDC\_expert}} \\
\hline
GPT-4o mini & 33.90 ± 0.38 & 15.18 ± 0.31 & 17.02 ± 0.27 & 1.25 ± 0.10 & 14.60 ± 0.25 & & 82.69 ± 0.35 & 12.65 ± 0.18 \\
LLaMa-3.1-8B & 31.17 ± 0.36 & 15.79 ± 0.32 & 16.43 ± 0.26 & 1.91 ± 0.11 & 13.91 ± 0.24 & & 82.18 ± 0.33 & 12.02 ± 0.17 \\
Mistral-7B & 32.12 ± 0.37 & \textbf{16.45 ± 0.33} & 16.48 ± 0.27 & 1.01 ± 0.09 & 14.13 ± 0.26 & & \textbf{82.77 ± 0.34} & \textbf{13.01 ± 0.19} \\
\textbf{Qwen2.5-7B} & \textbf{36.13 ± 0.39} & 15.67 ± 0.32 & \textbf{17.41 ± 0.28} & \textbf{4.16 ± 0.18} & \textbf{17.45 ± 0.30} & & 82.23 ± 0.34 & 12.17 ± 0.18 \\
\hline
\end{tabular}
\caption{Explanation performance comparison of various models across automatic evaluation metrics. Results are reported as mean ± standard deviation. The highest scores are in bold.}
\label{tab:exp_aut}
\end{table*}

To evaluate explanation quality, we rely on human assessments conducted by experts from the legal domain. We use three metrics: Clarity, Linking, and Usefulness, which together capture essential aspects of interpretability, legal alignment, and practical value. These metrics are particularly important in our setting, as the structured explanations produced by our framework differ significantly in form and granularity from the gold standard references in existing datasets \citep{nigam2024legal, malik2021ildc}, making reference-based automatic metrics less reliable.

Following \citep{nigam2024rethinking}, we adopt their definitions of Clarity and Linking, and introduce a new metric, Usefulness, as defined below:
\begin{itemize}
    \item Clarity: Measures how well-structured, readable, and logically coherent the explanation is.
    \item Linking: Assesses the degree to which the explanation offers a justifiable connection between the facts and the predicted outcome.
    \item Usefulness: Evaluates how informative and practically helpful the explanation is for a legal professional seeking to understand the court’s reasoning and apply it in real-world legal work.
\end{itemize}

For the evaluation, we recruited three legal experts, all practicing advocates with 7-8 years of professional experience. Each expert was provided with the case proceeding document, the predicted judgment outcome, and the corresponding explanation generated by the model. They rated 25 explanations per model across four LLMs, resulting in 100 annotated examples. Each explanation was rated independently on a 5-point Likert scale (1 = very poor, 5 = excellent) for all three criteria. Detailed guidelines provided to the evaluators are included in Appendix~\ref{appendix_guidelines}. 

Evaluating explanation quality across multiple random seeds was not performed due to the substantial human effort that would be required for assessing outputs from five separate runs. The evaluation was conducted on the explanation outputs of a single run. Inter-annotator agreement, measured using Fleiss' Kappa \cite{fleiss1971measuring}, was substantial (Clarity: 0.66, Linking: 0.70, Usefulness: 0.63). Table~\ref{tab:exp} reports the average scores per LLM per metric. Among the models, GPT-4o mini achieved the highest scores across all three dimensions, followed by Mistral-7B, which also demonstrated strong performance in Clarity and Linking. 

While our primary focus is on human evaluation due to its greater reliability, we also report standard reference-based automatic metrics (ROUGE-1, ROUGE-2, ROUGE-L, BLEU, METEOR, BERTScore, and BLANC) for completeness. We use the ground truth explanations available in the PredEx and ILDC\_expert datasets as reference texts. The results are provided in Table~\ref{tab:exp_aut}. Qwen2.5-7B records the highest scores on most automatic metrics, including ROUGE-1, ROUGE-L, BLEU, and METEOR. Although automatic metrics favor Qwen2.5-7B, human evaluation results favor GPT-4o-mini. This contrast highlights the limitations of automatic metrics in fully capturing the explanation quality.

\subsection{Ablation Study}
\label{sec:ablation}

To assess the necessity and contribution of each component within the Vichara framework, we conduct a systematic ablation study across four stages: (1) rhetorical role classification, (2) case context construction, (3) decision point extraction, and (4) present court ruling generation. 

\subsubsection{Experimental Setup}
We create a series of ablated variants of the full pipeline by selectively removing individual stages while holding all other configurations constant. When a stage is removed, subsequent prompts are minimally adjusted to maintain coherence and ensure functionality without referencing missing inputs. All evaluations are performed using the GPT-4o mini model on the ILDC\_expert dataset. We report accuracy and macro-F1 for the predictive component (judgment prediction), and Clarity, Linking, and Usefulness scores for the human evaluation of explanations. Judgment prediction results are reported as the mean of five independent runs with different random seeds, along with the standard deviation reflecting variation across seeds. The explanation evaluation was conducted on 15 explanation outputs from a single run, as assessing multiple runs would require substantial additional human effort. It was carried out by the same three legal experts who evaluated the explanations in Section~\ref{results:ex}. For explanation quality, we report the mean score across the three evaluators, with the standard deviation representing variation among evaluators.

\subsubsection{Ablation Configurations}

The ablation study consisted of five configurations:

\begin{enumerate}[noitemsep]
    \item \textbf{Full Vichara (All Stages)}: The complete pipeline as described in Section~\ref{methodology}.
    \item \textbf{Without Rhetorical Role Classification}: Sentence-level segmentation and rhetorical role labeling are omitted. The entire case proceeding document is passed as input for Case Context Construction (Section~\ref{context}). During Present Court Ruling Generation (Section~\ref{present}), the final statement—the last sentence originally labeled as `Ruling by Present Court'—is excluded from the input prompt.
    \item \textbf{Without Case Context Construction}: Case context is not generated and therefore not provided as input to the stages of Present Court Ruling Generation (Section~\ref{present}), Judgment Prediction (Section~\ref{pred}), or Structured Explanation Generation (Section~\ref{explanation}).
    
    \item \textbf{Without Decision Point Extraction}: The model directly generates the present court ruling from the input case proceeding document, the case context, and the final statement, without relying on decision points. Decision points are also not provided as input during Structured Explanation Generation (Section~\ref{explanation}).

    \item \textbf{Without Present Court Ruling Generation}: The present court ruling is not generated, and therefore not provided as input to the stages of Judgment Prediction (Section~\ref{pred}), or Structured Explanation Generation (Section~\ref{explanation}). Instead, the framework predicts the final judgment directly by comparing the case context with the extracted decision points. 

\end{enumerate}

\subsubsection{Results and Discussion}

\begin{table*}[t]
\centering
\small
\setlength{\tabcolsep}{5pt}
\begin{tabular}{lccccc}
\toprule
\textbf{Configuration} & \textbf{Accuracy} & \textbf{Macro F1} & \textbf{Clarity} & \textbf{Linking} & \textbf{Usefulness} \\
\midrule
\textbf{Full Vichara (All Stages)} & \textbf{80.36 $\pm$ 0.37} & \textbf{81.25 $\pm$ 0.34} & \textbf{4.57 $\pm$ 0.22} & \textbf{4.96 $\pm$ 0.21} & \textbf{4.37 $\pm$ 0.18} \\
Without Rhetorical Role Classification & 79.12 $\pm$ 0.45 & 78.85 $\pm$ 0.41 & 4.32 $\pm$ 0.28 & 4.66 $\pm$ 0.24 & 4.21 $\pm$ 0.25 \\
Without Case Context Construction & 77.84 $\pm$ 0.52 & 77.40 $\pm$ 0.47 & 4.21 $\pm$ 0.31 & 4.55 $\pm$ 0.26 & 4.18 $\pm$ 0.20 \\
Without Decision Point Extraction & 71.65 $\pm$ 0.61 & 70.92 $\pm$ 0.57 & 3.80 $\pm$ 0.35 & 4.10 $\pm$ 0.28 & 3.95 $\pm$ 0.33 \\
Without Present Court Ruling Generation & 73.22 $\pm$ 0.54 & 72.86 $\pm$ 0.49 & 3.95 $\pm$ 0.30 & 4.20 $\pm$ 0.27 & 4.00 $\pm$ 0.29 \\
\bottomrule
\end{tabular}
\caption{Ablation study results on the \textit{ILDC\_expert} dataset. Each configuration removes one stage from the Vichara pipeline. Reported values are mean $\pm$ standard deviation. Bold values indicate the highest performance for each metric.}
\label{tab:ablation}
\end{table*}

The results of the ablation study are presented in Table~\ref{tab:ablation}. Each stage of the Vichara pipeline demonstrably contributes to both predictive performance and explanation quality. The full configuration achieves the highest prediction performance, as well as top human evaluation scores for explanation quality. Removing individual components results in consistent declines across all metrics, confirming that the multi-stage architecture is essential for producing legally coherent and interpretable outcomes.
Eliminating Decision Point Extraction (Configuration 4) produces the largest drop in predictive performance, reducing macro-F1 by nearly 11 points. Eliminating Rhetorical Role Classification (Configuration 2) produces the smallest drop in predictive performance and explanation quality.

% Omitting Present Court Ruling Generation (Configuration 5) also significantly decreases performance. The removal of Case Context Construction (Configuration 3) and Rhetorical Role Classification (Configuration 2) leads to moderate declines in accuracy, macro-F1, and explanation quality, suggesting that these stages contribute to coherent and focused reasoning. 

\section{Conclusions}
\label{conclusions}

We introduced Vichara, a framework for appellate judgment prediction and explanation tailored to the Indian judicial system. By representing legal documents as sequences of decision points, Vichara enables accurate outcome prediction and interpretable explanation generation. Our structured explanation format, grounded in legal reasoning conventions, supports transparency and practical usability. Through experiments on two Indian legal datasets, PredEx and ILDC\_expert, we demonstrated that Vichara not only achieves strong performance across both proprietary and open-weight LLMs, but also surpasses existing benchmark results on these datasets for judgment prediction. Notably, smaller models such as Llama-3.1-8B, Qwen2.5-7B, and Mistral-7B, when used within Vichara, achieved results comparable to larger models like GPT-4o mini, offering a viable path for resource-constrained deployments. Human evaluation validated the quality of the generated explanations, showing high scores across Clarity, Linkage, and Usefulness. Vichara contributes a step forward in the development of explainable AI systems for judicial applications. 

Future work will explore methods to reduce computational overhead and adapt the framework to other case types and legal jurisdictions. 

\section{Limitations}
\label{limitations}
While Vichara demonstrates strong performance in both judgment prediction and structured explanation generation, several limitations remain.

Our evaluation is restricted to the Indian judiciary, with a primary focus on appellate-level cases from the Supreme Court and selected High Courts. Although the core methodology may be applicable to other legal systems, transferring the approach to jurisdictions with different procedural structures, legal doctrines, or language conventions would require substantial adaptation and validation.

Vichara relies on prompt-based querying of LLMs at multiple stages of the pipeline. This approach introduces variability due to the non-deterministic nature of model outputs and the sensitivity of results to prompt phrasing.

The human evaluation of explanation quality was conducted on a limited sample consisting of 25 explanations per language model, reviewed by three advocates. While this provides useful qualitative insights, broader evaluation involving a more diverse pool of legal professionals and a wider range of case types is necessary to establish generalizability and practical relevance.

Finally, the multi-stage architecture of Vichara, which includes multiple LLM calls, may impose computational and deployment challenges. These constraints could be particularly limiting in environments with restricted resources or strict latency requirements.

Future research will focus on investigating methods to reduce computational overhead and expanding evaluation to include additional legal domains and jurisdictions. This includes exploring strategies such as prompt optimization and model distillation to improve scalability while maintaining performance in real-world legal applications. To broaden evaluation, we plan to assess the framework on a wider range of case types, court levels, and jurisdiction-specific datasets, enabling a more comprehensive understanding of its effectiveness across diverse legal contexts.

\section*{Ethical Statement}

In conducting this research, we adhered to ethical standards in both data usage and human evaluation. All legal case documents used in our experiments are publicly available and drawn from established open-access legal datasets. No private, confidential, or sensitive information was accessed or utilized at any stage. For the human evaluation of model-generated explanations, we engaged three practicing advocates with formal legal training and courtroom experience. Their participation was entirely voluntary. The evaluation was conducted with informed consent, and participants were briefed on the purpose and scope of the study.

\bibliography{aaai2026}

\appendix
\section{Human Evaluation Guidelines}
\label{appendix_guidelines}

The following instructions were provided to human evaluators for assessing the quality of the generated explanations. Each explanation was evaluated independently along three dimensions using a 5-point Likert scale.

\subsection{Clarity}

Measures how clearly the explanation is written and how well it communicates the rationale behind the decision.

\begin{itemize}
    \item[1:] The explanation is confusing or incoherent. The rationale is difficult to follow.
    \item[2:] Some parts are understandable, but the reasoning is vague or underdeveloped.
    \item[3:] The explanation is moderately clear but may lack smooth flow or sufficient detail.
    \item[4:] The rationale is clearly presented and easy to follow. Terminology is appropriate.
    \item[5:] The explanation is very well-written, logically structured, and highly understandable.
\end{itemize}

\subsection{Linking}

Measures how well the explanation connects the facts and legal reasoning to the final predicted outcome (e.g., Appeal Granted or Dismissed).

\begin{itemize}
    \item[1:] The explanation does not connect to the outcome at all or is highly inconsistent.
    \item[2:] Weak or unclear linkage between the reasoning and the final decision.
    \item[3:] Some linkage exists, but gaps or ambiguities are present.
    \item[4:] Clear and logical connection to the judgment, with minimal gaps.
    \item[5:] Strong and coherent justification that clearly supports the predicted decision.
\end{itemize}

\subsection{Usefulness}

Measures how useful the explanation would be for a human reader, particularly a legal practitioner, trying to understand the reasoning behind the AI’s decision.

\begin{itemize}
    \item[1:] The explanation is not helpful or usable in any practical sense.
    \item[2:] Limited usefulness; lacks essential detail or context.
    \item[3:] Somewhat helpful, but may miss key points or feel generic.
    \item[4:] Offers clear value in understanding the reasoning and potential legal implications.
    \item[5:] Highly informative and usable; effectively mirrors real-world legal reasoning.
\end{itemize}

\section{Prompts for Each Processing Step}
\begin{figure*}[htbp]
  \centering
  \includegraphics[width=0.9\linewidth]{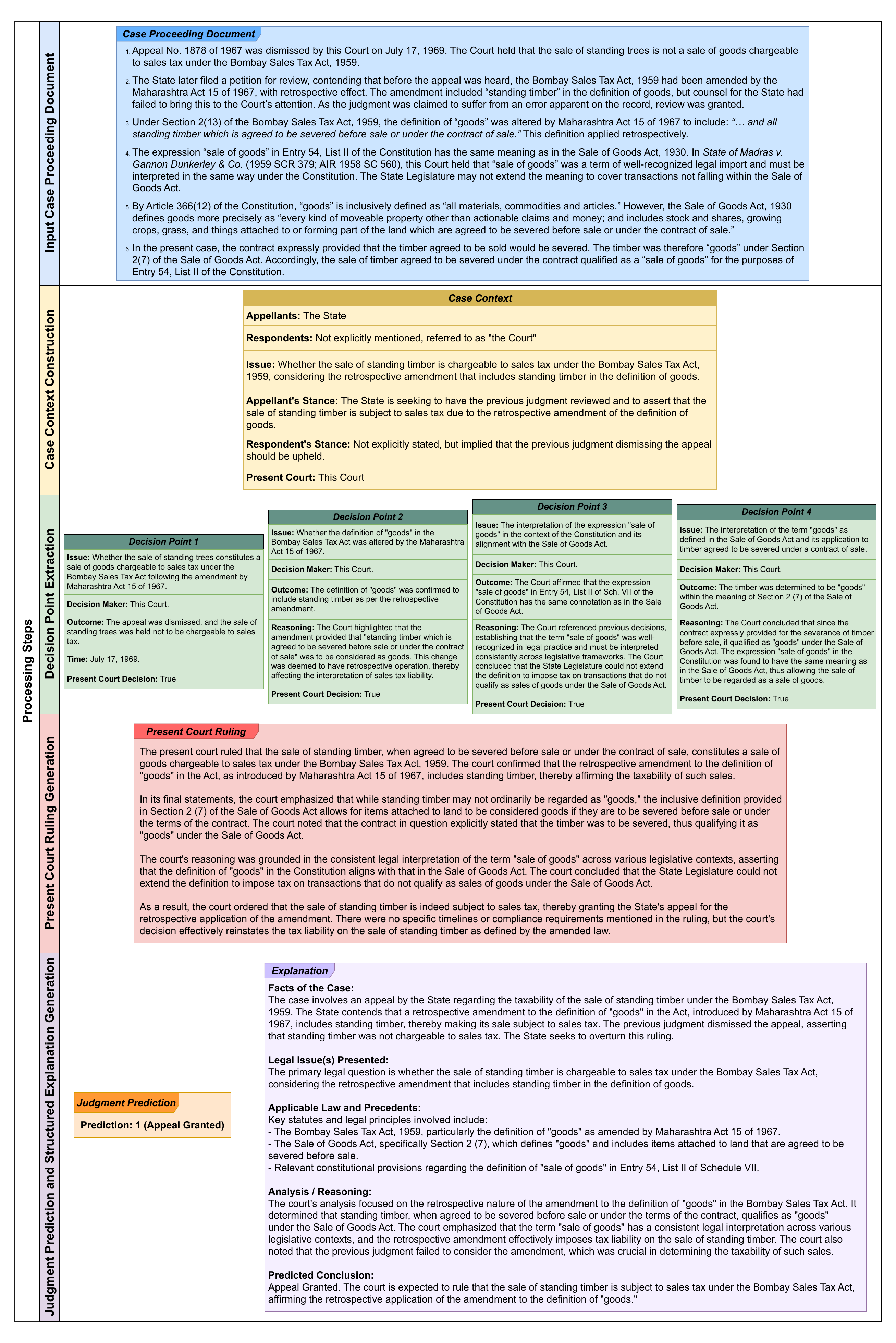} \hfill
  \caption {Outputs at each processing step for an example court case.}
  \label{example}
\end{figure*}

Tables \ref{tab:prompts1}, \ref{tab:prompts2}, and \ref{tab:prompts3} contain the prompts used for each stage of the processing pipeline. The prompts were developed through an iterative prompt engineering process, inspired by the principles on prompt design \citep{phoenix2024prompt}. Figure~\ref{example} illustrates the outputs generated at each processing step using GPT-4o mini, for an example court case.

\newpage
\clearpage
\begin{table*}[t]
\centering
\scriptsize
\begin{tabular}{|p{0.07\textwidth}|p{0.9\textwidth}|}
\hline
\textbf{Step} & \textbf{Prompt} \\
\hline

Case Context Construction &

You are a legal assistant helping summarize appeal case details. \newline

Given the following facts from an appeal case document, extract the following information about the current appeal only:\newline

1. Appellants – the persons or entities filing the current appeal. If their name is not mentioned, write what they are referred to as in the text (e.g., "the petitioner", "the appellant").\newline
2. Respondents – the persons or entities against whom the current appeal is filed. If their name is not mentioned, write what they are referred to as in the text (e.g., "respondent 1", "the respondent-Management").\newline
3. Issue – the main legal or factual issue being disputed in the current appeal.\newline
4. Appellant's Stance (in the current appeal) – clearly state what the appellant is arguing for or seeking in the present appeal.\newline
5. Respondent's Stance (in the current appeal) – clearly state what the respondent is arguing for or seeking in the present appeal.\newline
6. Present Court – the court deciding the present appeal (e.g., Supreme Court of India, High Court of Bombay).\newline

\verb|###| Important instructions:\newline

- Do NOT assume the appellant is the party introduced first. Carefully check who has filed the current appeal.\newline
- Do NOT summarize or include opinions or findings of lower courts unless those are being specifically challenged in this appeal.\newline
- Focus on the actual parties to the legal dispute.\newline
- Do NOT invent names or facts. If something is not mentioned, leave it as an empty string.\newline
- Use only the output format specified below.\newline

\verb|###| Output Format:\newline

\{\{ \newline"appellants": "<name or description of appellant>", \newline"respondents": "<name or description of respondent>",\newline "issue": "<brief summary of the legal/factual issue>", \newline"appellant\_stance": "<stance of the appellant>", \newline"respondent\_stance": "<stance of the respondent>", \newline"present\_court": "<name of the court currently deciding the appeal>" \newline\}\}\newline

\verb|###| Facts: <\{facts\}> \\

\hline

Decision Point Extraction &

You are a legal assistant tasked with extracting **all decision points** from an excerpt of a court case proceeding. Decision points are discrete legal determinations made at different stages throughout the case. \newline
 
Given the following text and the present court, extract **all identifiable decision points** and output them in strict JSON format as a list of objects.\newline
 
Each decision point object should include:\newline
- "issue": the legal issue or question being addressed\newline
- "decision\_maker": the court or authority that made the decision (e.g., Supreme Court, Trial Court, High Court)\newline
- "outcome": the result or resolution of the issue\newline
- "time": (optional) the date or timeframe of the decision if mentioned\newline
- "reasoning": (optional) summary of the Court's reasoning, including references to statutes, arguments, facts, or precedents\newline
- "present\_court\_decision": true if the "decision\_maker" is the same as the present court provided, otherwise false\newline

\verb|###| Important instructions:\newline
- Do not include any extra text or explanation.\newline
- Do not assume or hallucinate decision makers.\newline
- Do NOT include triple backticks (```).\newline

\verb|###| Output Format:\newline

[\{\newline "issue": "<string>", \newline"decision\_maker": "<string>",\newline "outcome": "<string>",\newline "time": "<string or null>", \newline"reasoning": "<string or null>", \newline"present\_court\_decision": <true or false> \newline\}]\newline

\verb|###| Input: \newline
Present Court: "<\{present\_court\}>"\newline
Text: <\{group\_text\}> \\

\hline
\end{tabular}
\caption{Prompts for Case Context Construction and Decision Point Extraction}
\label{tab:prompts1}
\end{table*}

\begin{table*}[t]
\centering
\scriptsize
\begin{tabular}{|p{0.07\textwidth}|p{0.9\textwidth}|}
\hline
\textbf{Step} & \textbf{Prompt} \\
\hline

Present Court Ruling Generation &

Your goal is to identify the **final ruling of the present court** in this appeal — that is, what the present court ultimately decided and ordered.\newline

\verb|###| Case Context:\newline
 
Below is the context of the case, which clearly identifies:
- Who the **appellant** is (the party who filed the appeal) \newline
- Who the **respondent** is (the party defending against the appeal) \newline
- What the **main issue** of the appeal is. \newline
- Appellant's Stance (in the current appeal) – What the appellant is arguing for or seeking **in the present appeal** \newline
- Respondent's Stance (in the current appeal) – What the respondent is arguing for or seeking **in the present appeal** \newline
 
Please pay close attention to this information, it overrides any assumptions you might make from the decision points. If the appellants, respondents or the issue of the appeal are not mentioned in the context, ONLY then infer these from the decision points. \newline
 
<\{context\}>\newline
---

\verb|###| Decision Points:\newline
 
Decision points are key legal determinations made at different stages throughout the case. The below decision points collectively summarize the key determinations the present court made throughout the case. \newline
 
<\{present\_court\_points\}>\newline
---

\verb|###| Final Statements from the Present Court:\newline
 
This section contains the last official statements or conclusions made by the present court in this appeal. These are the most authoritative and conclusive indication of the court’s final position and must be treated as such. \newline
 
<\{final\_statement\}>\newline
---
 
\verb|###| Your Task:\newline
 
1. Provide a comprehensive explanation of the final ruling, including:\newline
   - Specific reliefs granted or denied\newline
   - Any orders or directions issued\newline
   - The court’s reasoning and key factors considered\newline
   - Relevant timelines or compliance expectations\newline
2. Focus on the decision points to determine what the present court considered during this appeal.\newline 
3. Use the **Final Statements from the Present Court** to determine what the court ultimately ruled.\newline
---
 
\verb|###| Important:\newline

- Do not confuse appellants and respondents. Use the parties as stated in the case context.\newline
- Respond clearly and concisely.\newline
- Do NOT state whether the appeal was granted or dismissed. Your response should only describe the final ruling of the present court.\newline
---

\verb|###| Output Format:\newline

Final Ruling:\newline
<Provide a detailed and comprehensive explanation of the present court’s final decision in the appeal.>\newline
\\ \hline
\end{tabular}
\caption{Prompt for Present Court Ruling Generation}
\label{tab:prompts2}
\end{table*}

\begin{table*}[t]
\centering
\scriptsize
\begin{tabular}{|p{0.07\textwidth}|p{0.9\textwidth}|}
\hline
\textbf{Step} & \textbf{Prompt} \\
\hline

Judgment Prediction &

You are a legal assistant helping to analyze the outcome of an appeal. Your task is to determine whether the **present court’s final ruling** aligns with what the **appellant** was seeking in this appeal.\newline

\verb|###| Case Context:\newline
The context below includes: \newline
- Appellants: The **appellant** (party who filed the appeal) \newline
- Respondents: The **respondent** (party defending the appeal) \newline
- Issue: The **main issue** \newline
- Appellant's Stance: What the **appellant is seeking** in the current appeal \newline
- Respondent's Stance: What the **respondent is seeking** in the current appeal \newline
- Present Court: The court deciding the present appeal. \newline
 
<\{context\}>\newline
---
 
\verb|###| Final Court Ruling:\newline
<\{court\_ruling\}>\newline
---
 
\verb|###| Your Task:\newline
- If the **court fully or partially granted what the appellant was seeking**, output: `Prediction: 1`\newline
- If the **court did not grant what the appellant was seeking**, output: `Prediction: 0`\newline
---

\verb|###| Output Format:\newline
Prediction: <0 or 1> \\

\hline

Structured Explanation Generation &

You are a legal assistant. Your task is to generate a structured legal explanation for the court's predicted decision in this appeal case.\newline

You are given:\newline
- The **case context** including appellant, respondent, issue, stances, and the court deciding the appeal\newline
- The **final court ruling** from the present court\newline
- A set of **decision points** extracted from the case. A **decision point** refers to a key moment in the case where a specific issue was considered, a responsible authority or decision-maker evaluated it, and a determination or outcome was reached.\newline
- The **predicted outcome** of the appeal\newline

Generate a structured explanation with the following sections:\newline
 
---\newline
Facts of the Case:\newline
[A brief summary of the background, parties involved, and what led to the appeal.]\newline

Legal Issue(s) Presented:\newline
[The legal question(s) the court had to decide.]\newline

Applicable Law and Precedents:\newline
[Key statutes, constitutional provisions, or case law relied on.]\newline

Analysis / Reasoning:\newline
[A logical application of law to facts, showing why the court ruled the way it did.]\newline

Predicted Conclusion:\newline
[Restate the predicted outcome using legal terminology (e.g., 'Appeal Allowed' or 'Dismissed').]\newline
---\newline
 
\verb|###| Case Context:\newline
<\{context\}>\newline

\verb|###| Final Court Ruling:\newline
<\{court\_ruling\}>\newline

\verb|###| Decision Points:\newline
<\{decision\_points\_text\}>\newline

\verb|###| Predicted Outcome:\newline
<\{predicted\_outcome\}>\newline
\\
\hline
\end{tabular}
\caption{Prompts for Judgment Prediction and Structured Explanation Generation}
\label{tab:prompts3}
\end{table*}

\clearpage

\end{document}